\PassOptionsToPackage{table}{xcolor}
\documentclass[10pt,twocolumn,letterpaper]{article}
\usepackage[pagenumbers]{cvpr} %

\newcommand{\mypar}[1]{\noindent\textbf{#1}}

\usepackage{duckuments}
\usepackage{comment}
\usepackage{xspace}

\usepackage{tabularx}
\usepackage{adjustbox}
\usepackage{multicol}
\usepackage{multirow}
\usepackage{caption}

\usepackage{subcaption}

\newcommand{\ours}{TreeMort-3T-UNet\xspace}

\usepackage{pifont}

\usepackage[table]{xcolor}
\definecolor{lncolor}{HTML}{FEE4C4}
\definecolor{baseline}{HTML}{E0E4E8}
\definecolor{ppink}{rgb}{0.98, 0.575, 0.89}

\newcommand\blfootnote[1]{%
  \begingroup
  \renewcommand\thefootnote{}\footnote{#1}%
  \addtocounter{footnote}{-1}%
  \endgroup
}

\definecolor{cvprblue}{rgb}{0.21,0.49,0.74}
\usepackage[pagebackref,breaklinks,colorlinks,allcolors=cvprblue]{hyperref}

\def\paperID{16413} %
\def\confName{CVPR}
\def\confYear{2025}

\title{Dual-Task Learning for Dead Tree Detection and Segmentation with Hybrid Self-Attention U-Nets in Aerial Imagery}

\author{
Anis Ur Rahman\textsuperscript{*}
\quad 
Einari Heinaro\textsuperscript{}
\quad
Mete Ahishali\textsuperscript{}
\quad
Samuli Junttila\textsuperscript{} \\ 
\small
\textsuperscript{}School of Forest Sciences, University of Eastern Finland, Joensuu 80101, Finland
}

\NewDocumentCommand\tc{O{red} m}{\textcolor{#1}{TODO. 
 #2}}

\begin{document}
\maketitle

\begin{abstract}
Mapping standing dead trees is critical for assessing forest health, monitoring biodiversity, and mitigating wildfire risks, for which aerial imagery has proven useful. However, dense canopy structures, spectral overlaps between living and dead vegetation, and over-segmentation errors limit the reliability of existing methods. This study introduces a hybrid postprocessing framework that refines deep learning-based tree segmentation by integrating watershed algorithms with adaptive filtering, enhancing boundary delineation, and reducing false positives in complex forest environments. Tested on high-resolution aerial imagery from boreal forests, the framework improved instance-level segmentation accuracy by 41.5\% and reduced positional errors by 57\%, demonstrating robust performance in densely vegetated regions. By balancing detection accuracy and over-segmentation artifacts, the method enabled the precise identification of individual dead trees, which is critical for ecological monitoring. The framework’s computational efficiency supports scalable applications, such as wall-to-wall tree mortality mapping over large geographic regions using aerial or satellite imagery. These capabilities directly benefit wildfire risk assessment (identifying fuel accumulations), carbon stock estimation (tracking emissions from decaying biomass), and precision forestry (targeting salvage loggings). By bridging advanced remote sensing techniques with practical forest management needs, this work advances tools for large-scale ecological conservation and climate resilience planning.
\blfootnote{\textsuperscript{*} Corresponding author: \texttt{anis.rahman@uef.fi}. \\ Code at \url{https://github.com/Global-Ecosystem-Health-Observatory/TreeMort}}.
\end{abstract}

% % Note that keywords are not normally used for peerreview papers.
% \begin{IEEEkeywords}
% Dead tree segmentation, forest mortality mapping, aerial imagery analysis, Self-Attention U-Net, high-resolution imagery.    
% \end{IEEEkeywords}

\section{Introduction}
\label{sec:intro}

Tree mortality serves as a critical indicator of forest health, with profound implications for carbon emissions, wildfire risks, and biodiversity conservation~\cite{parmesan2022climate}. The detection of dead trees is essential for tracking carbon emissions from decaying biomass, assessing wildfire fuel loads, and monitoring habitat loss amid intensifying climate change pressures, as highlighted by a comprehensive 2022 report on climate impacts~\cite{parmesan2022climate}. Accurate mapping of individual dead trees supports global initiatives, such as the UN Sustainable Development Goals, by informing reforestation strategies and biodiversity conservation efforts. However, identifying and delineating dead trees in remote sensing imagery remains a complex challenge due to dense canopy structures, spectral overlaps between living and dead vegetation, and over-segmentation errors that constrain traditional methods~\cite{weinstein2021benchmark}. These difficulties are compounded by spectral similarities, particularly in disturbed or seasonal forest ecosystems~\cite{international2025towards}.

Aerial imagery from drones or aircraft provides high-resolution views of forest landscapes, while multispectral data---including near-infrared (NIR) bands---reveal vegetation health invisible to the naked eye. Within this context, instance segmentation has emerged as a powerful tool, enabling pixel-level classification of tree canopies and the separation of individual trees. This capability is vital for accurately assessing tree mortality and its ecological impacts, especially in dense forest environments where canopies overlap. Recent studies on advanced deep learning models demonstrate their effectiveness in distinguishing individual trees~\cite{ji2022automated}, while a 2021 benchmark study underscores persistent challenges such as dense canopies, spectral overlaps, and segmentation errors in aerial imagery analysis~\cite{weinstein2021benchmark}. Despite its promise, instance segmentation in remote sensing faces significant hurdles. Existing datasets often lack spectral richness (e.g., NIR bands) and geographic diversity, limiting their ability to capture the variability in dead tree appearances across ecosystems~\cite{weinstein2021benchmark,ma2019deep}. Moreover, pre-trained deep learning models, typically developed for natural images, struggle to generalize to high-resolution aerial data due to differences in scale, perspective, and spectral composition, as noted in a 2020 meta-analysis~\cite{ma2019deep}. These limitations impede reliable forest monitoring, particularly over large regions where computational efficiency is paramount.

Recent advances have sought to address these challenges by integrating multi-modal data and task-specific postprocessing techniques~\cite{wang2024individual, zhou2021qualification}. Combining RGB and NIR imagery enhances spectral discrimination, improving detection accuracy, as shown in a 2021 study on multispectral UAV imagery for forest analysis~\cite{zhou2021qualification}. Meanwhile, methods like watershed-based refinement, explored in a 2022 study, enhance the separation of overlapping objects in dense forests~\cite{wang2024individual}. However, existing approaches often struggle to balance detection accuracy with over-segmentation artifacts, limiting their practical utility for operational ecological monitoring. This study introduces a novel hybrid framework that surpasses these methods by integrating a Self-Attention U-Net with watershed-guided postprocessing. Unlike prior deep learning models that falter in dense canopies~\cite{ji2022automated}, this approach leverages centroid localization and adaptive boundary refinement to achieve superior instance-level accuracy in boreal forest settings, where mapping tree mortality is crucial for wildfire risk assessment and carbon stock estimation.

The key contributions of this work are threefold:

\begin{enumerate}
\item A unified architecture that simultaneously performs instance segmentation and centroid localization using a domain-adapted ResNet-34 encoder, optimized for handling overlapping objects and multispectral variability in aerial imagery.
\item A novel combination of Binary Cross-Entropy, Dice, Focal, and Mean Squared Error losses that robustly balances pixel-level segmentation and precise centroid localization while mitigating class imbalance.
\item A multi-step refinement process incorporating adaptive thresholding, centroid-guided enhancement, morphological operations, and watershed segmentation to accurately delineate individual tree instances.
\end{enumerate}

This paper presents a comprehensive framework that improves dead tree segmentation accuracy by harmonizing deep learning with watershed-based refinement, offering a scalable solution for large-scale ecological monitoring. The remainder of the paper is organized as follows: Section~\ref{sec:related} reviews related work in remote sensing instance segmentation; Section~\ref{sec:method} details the proposed methodology; Section~\ref{sec:results} presents experimental results; and Section~\ref{sec:conclusions} concludes with implications and future directions.

\section{Related Work}
\label{sec:related}

\mypar{Datasets for instance segmentation in remote sensing.} The development of instance segmentation methods for remote sensing, particularly for ecological applications such as dead tree detection, relies on datasets that capture the spectral, spatial, and geographic intricacies of aerial imagery. General-purpose datasets like COCO-Stuff~\cite{caesar2018coco} and OpenImages~\cite{kuznetsova20openimages} are fundamental for natural image analysis but lack the spectral richness and high spatial resolution required for remote sensing applications. Specialized datasets such as TreeSatAI~\cite{ahlswede2022treesatai}, ISPRS Semantic Labeling~\cite{rottensteiner2014isprs}, and TreeSeg~\cite{speckenwirth2024treeseg} provide high-resolution aerial annotations tailored to vegetation analysis, but they are often geographically biased toward temperate forests or urban environments. The U-Pixels dataset~\cite{ramos2025leveraging} attempts to improve vegetation classification through multi-modal data, while the Fallen Tree Segmentation dataset~\cite{polewski2021instance} extends segmentation capabilities to downed trees in aerial color-infrared imagery. Despite these advancements, many datasets lack RGB-NIR integration, which is crucial for distinguishing dead trees across different ecosystems, and rely on polygonal annotations that approximate instance boundaries without providing precise centroid markers.

\mypar{Deep learning advancements in remote sensing instance segmentation.} Deep learning has significantly improved instance segmentation for remote sensing, with architectures such as Mask R-CNN~\cite{he2017mask} and U-Net~\cite{ronneberger2015u} serving as benchmarks for object detection and pixel-wise segmentation. However, aerial imagery presents unique challenges, including multi-scale feature extraction to accommodate high-resolution data, spectral fusion to differentiate vegetation states, and precise separation of overlapping instances like tree canopies. To tackle these issues, models such as PSPNet~\cite{zhao2017pyramid} and attention-based mechanisms~\cite{hu2018squeeze} enhance feature extraction by emphasizing discriminative spectral-spatial cues, while vision transformers~\cite{alexey2020image} improve contextual understanding of high-dimensional multi-spectral data, although their computational demands limit scalability. Recent adaptations have focused on integrating spectral information more effectively, with RGB-NIR fusion techniques~\cite{lv2023pruning} improving vegetation segmentation and LiDAR-RGB fusion~\cite{weinstein2019individual} enhancing the structural delineation of tree crowns. Lightweight architectures, including MobileNet~\cite{howard2017mobilenets}, strike a balance between computational efficiency and accuracy, making large-scale remote sensing applications more feasible, while graph-based networks~\cite{jiang2024heterogeneous} model spatiotemporal relationships in satellite imagery to better capture environmental dynamics. Despite these advancements, domain gaps between pre-trained models and aerial data hinder generalization~\cite{zhu2017deep}, and the high computational cost of processing large, high-resolution datasets restricts real-time applicability. Hybrid strategies that combine deep learning with domain-specific knowledge, such as spectral indices and geometric priors, have been proposed to mitigate these challenges, while instance segmentation frameworks continue to evolve through loss function optimizations~\cite{azad2023loss}, including focal loss, Dice loss, and hybrid cost functions tailored for remote sensing applications.

\mypar{Centroid-based approaches in remote sensing.} Centroid-based instance segmentation has gained traction as an effective strategy for remote sensing applications requiring precise object localization. Centroid-UNet~\cite{deshapriya2021centroid} employs a UNet-inspired architecture to generate centroid maps for object detection in aerial images. At the same time, CentroidNet~\cite{dijkstra2019centroidnet} integrates object localization and counting within a deep learning framework, and CentroidNetV2~\cite{dijkstra2021centroidnetv2} refines small-object segmentation through hybrid deep learning techniques. In infrared small target detection, UCDNet~\cite{xu2023ucdnet} utilizes a double U-shaped network to cascade segmentation and centroid map prediction, demonstrating the effectiveness of dual-task learning. Despite their advantages, these models often struggle with overlapping tree canopies and spectral variability in remote sensing data. Watershed-based postprocessing techniques~\cite{wang2024individual} have been introduced to refine tree crown delineation from UAV imagery, improving object separation in dense forests while maintaining computational efficiency.

\mypar{Multi-modal data integration for enhanced segmentation accuracy.} Multi-modal data integration has become crucial for improving instance segmentation accuracy in remote sensing, particularly for dead tree detection. Traditional RGB-based methods often fail to distinguish spectrally similar objects, such as dead trees and senescent vegetation. The incorporation of NIR bands~\cite{lv2023pruning} enhances segmentation by leveraging spectral contrasts, as healthy foliage reflects strongly NIR, while dead vegetation exhibits reduced reflectance. LiDAR-RGB fusion techniques~\cite{weinstein2019individual} provide additional structural information, facilitating the separation of overlapping tree canopies. Vision transformers have expanded multi-modal fusion capabilities by processing heterogeneous data sources through unified attention mechanisms, with multi-scale vision transformers~\cite{gu4762408multi} demonstrating success in segmenting Arctic permafrost by jointly analyzing spectral and thermal data. However, challenges such as spatial misalignment across modalities and sensor noise remain prevalent. UAV-based multispectral classification models~\cite{zhou2021qualification} offer an alternative approach, improving segmentation accuracy in mixed forest systems by leveraging deep learning to enhance spectral fusion and feature differentiation.

\mypar{Summary.} Instance segmentation in remote sensing has progressed through advancements in datasets, deep learning architectures, centroid-based methods, and multi-modal fusion. While datasets such as TreeSatAI, ISPRS Semantic Labeling, and TreeSeg provide valuable resources, their geographic biases, limited multi-modal integration, and lack of precise localization data constrain their applicability. Deep learning-based segmentation, including multi-scale networks, transformer models, and optimized loss functions, has improved feature extraction but still faces challenges in spectral variability and computational scalability. Centroid-based approaches such as Centroid-UNet, CentroidNet, and UCDNet offer promising solutions for precise localization, yet further adaptation is required to handle overlapping canopies in aerial imagery. Multi-modal fusion using RGB-NIR and LiDAR-RGB techniques has demonstrated improvements in segmentation accuracy, though challenges related to feature alignment and computational efficiency persist. Future research must develop robust, scalable methods integrating spectral, structural, and spatial information into a unified segmentation framework.

\section{Methodology}
\label{sec:method}

\subsection{\ours{} Architecture for Multi-Task Learning}

The \ours{} architecture addresses the challenges of instance segmentation and instance localization in high-resolution remote sensing imagery. To capture the complex spatial and spectral characteristics of aerial data, the architecture employs a three-headed decoder that simultaneously predicts (i)~a segmentation mask, (ii)~a centroid localization map, and (iii)~a hybrid SDT-boundary map. These outputs are fused in a subsequent postprocessing stage to achieve precise instance delineation.

\begin{figure*}[t!]
\centering
\includegraphics[width=1\linewidth]{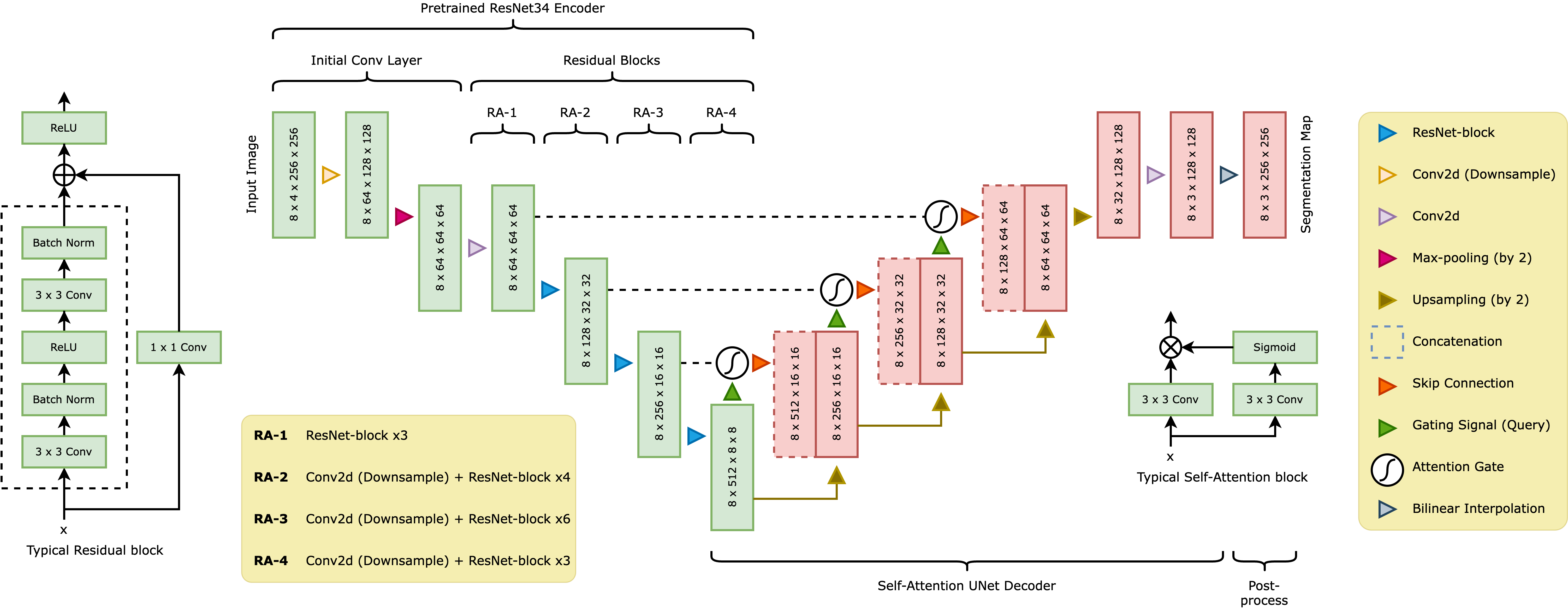}
\caption{The \ours{} architecture for multi-task learning in instance segmentation. A pre-trained ResNet-34 encoder extracts robust features from multispectral (RGB-NIR) aerial imagery. The hierarchical decoder, connected via skip connections, bifurcates into three output branches: a segmentation head that produces a binary mask, a centroid head that outputs a Gaussian heatmap for instance localization, and a hybrid head that generates a signed distance transform (SDT) combined with explicit boundary cues. Self-Attention modules refine feature maps to handle overlapping and densely clustered tree canopies effectively.}
\label{fig:arch}
\vspace{-1em}
\end{figure*}

\mypar{Encoder with transfer learning from the FLAIR-INC dataset.} The encoder leverages a ResNet-34 backbone pre-trained on ImageNet and fine-tuned on the FLAIR-INC dataset. This dataset, composed of RGB-NIR imagery from the BD ORTHO aerial product, spans 75 radiometric domains and diverse land cover classes. The pre-training confers three primary benefits:
\begin{enumerate}
\item \textbf{Domain Robustness:} The diversity of the FLAIR-INC dataset ensures resilience against domain shifts, enabling robust performance across varying spatial and temporal contexts.
\item \textbf{Spectral Sensitivity:} The inclusion of the NIR channel enhances the encoder’s capacity to distinguish subtle vegetation characteristics.
\item \textbf{Scalability:} High spatial resolution (0.2 m/pixel) and geographic diversity allow the encoder to generalize well to new datasets with similar imaging properties.
\end{enumerate}

Mathematically, an input image $\mathbf{X} \in \mathbb{R}^{H \times W \times 4}$ (RGB-NIR) is processed by the encoder to yield a latent representation:
\begin{align}
\label{eq:latent}
\mathbf{Z} = \texttt{Encoder}(\mathbf{X}), \quad \mathbf{Z} \in \mathbb{R}^{h \times w \times d}.
\end{align}
Here, $\mathbf{Z}$ encapsulates the essential spatial and spectral features---such as textures, edges, and contextual cues---that are crucial for distinguishing tree canopies from the background. This rich feature map is then passed to a multi-headed decoder, which exploits these learned representations to simultaneously generate a segmentation mask, a centroid heatmap, and a hybrid SDT-boundary map. In this way, $\mathbf{Z}$ forms the foundational basis for all subsequent multi-task predictions.

\mypar{Decoder with triple-output heads.} The decoder is augmented with three distinct output branches:

\begin{enumerate}
    \item \textbf{Segmentation Head:} This branch predicts a binary mask $\mathbf{Y}_s \in {0, 1}^{H \times W}$ that distinguishes dead tree canopies from the background. A sigmoid activation is applied to the logits to generate probability maps.

    \item \textbf{Centroid Localization Head:} The centroid branch outputs a Gaussian heatmap $\mathbf{Y}_c \in \mathbb{R}^{H \times W}$, where each peak corresponds to the center of a dead tree canopy. The Gaussian kernel is applied to ground-truth centroids to produce smooth, localized intensity maps.

    \item \textbf{Hybrid SDT-Boundary Head:} The new hybrid branch produces a map $\mathbf{Y}_h \in \mathbb{R}^{H \times W}$ that encodes a Signed Distance Transform (SDT) fused with explicit boundary cues. In this representation, boundary pixels are ideally assigned values near $–1$, while nonboundary pixels represent normalized distances up to $+1$. This branch is critical for refining object boundaries, especially in regions with overlapping or densely clustered canopies.
\end{enumerate}

Skip connections between the encoder and decoder facilitate the preservation of fine spatial details. Despite being trained without additional activations, the centroid and hybrid branches complement the segmentation head by providing precise localization and boundary refinement cues during postprocessing.

\mypar{Attention mechanisms.} Attention modules are integrated into the decoder to enhance relevant feature maps and suppress background noise dynamically. This mechanism is particularly important for multi-spectral imagery, as it aids in focusing on subtle spectral and spatial variations critical for distinguishing dead tree canopies.

\mypar{Rationale behind \ours{}.} \ours{} is engineered to overcome the challenges of high-resolution aerial imagery, including overlapping canopies, class imbalance, and spectral variability. Key innovations include:
\begin{itemize}
    \item A ResNet-34 encoder fine-tuned on the diverse FLAIR-INC dataset, ensuring robustness against domain shifts.
    \item A triple-output decoder that consolidates segmentation, centroid localization, and boundary delineation within a single unified model.
    \item The generation of Gaussian heatmaps and hybrid SDT-boundary maps that facilitate instance separation in complex scenes.
    \item Attention mechanisms that refine feature extraction, leading to improved segmentation accuracy even in cluttered environments.
\end{itemize}

\mypar{Visual summary.} Figure~\ref{fig:arch} illustrates the overall architecture, detailing the flow from input image through feature extraction, multi-task decoding, and final output generation. This diagram highlights the interplay between segmentation, centroid, and hybrid outputs, providing a conceptual framework for the subsequent postprocessing pipeline.

\subsection{Hybrid Loss Function for Multi-Task Training}

The hybrid loss function is specifically tailored to jointly optimize the three outputs. It integrates several complementary loss components:

\mypar{Segmentation loss.} A combination of weighted Binary Cross-Entropy (BCE) loss and Dice loss drives the accurate separation of dead tree canopies from the background:
\begin{align}
\label{eq:dice}
\mathcal{L}_{\texttt{seg}} = \mathcal{L}_{\texttt{BCE}} + \lambda_{\texttt{Dice}} \mathcal{L}_{\texttt{Dice}}.
\end{align}

\mypar{Centroid localization loss.} Mean Squared Error (MSE) loss is used to penalize deviations between the predicted Gaussian heatmap and the ground truth:
\begin{align}
\label{eq:loss_centroid}
\mathcal{L}_{\texttt{centroid}} = \frac{1}{N} \sum{i=1}^{N} \left( P^c_i - Y^c_i \right)^2.
\end{align}

\mypar{Hybrid SDT-boundary loss.} The hybrid branch is optimized using a regression loss that directly compares the predicted SDT-boundary map to the ground truth. The loss is computed in two regimes:
\begin{itemize}
    \item \textbf{SDT Loss:} For non-boundary pixels, a Smooth L1 loss ensures the predicted distance values match the normalized ground truth.
    \item \textbf{Boundary Loss:} For boundary pixels (with ground truth $–1$), an L1 loss directly penalizes deviations from $–1$. Due to the extreme rarity of boundary pixels, this loss is up-weighted to ensure its influence during training.
\end{itemize}

The overall hybrid loss is given by:
\begin{align}
\label{eq:loss_hybrid}
\mathcal{L}_{\texttt{hybrid}} = \lambda_{\texttt{SDT}} \mathcal{L}_{\texttt{SDT}} + \lambda_{\texttt{boundary}} \mathcal{L}_{\texttt{boundary}},
\end{align}
where $\lambda{\texttt{SDT}}$ and $\lambda_{\texttt{boundary}}$ are hyperparameters determined empirically.

\mypar{Total loss.} The combined loss function is a weighted sum of the individual components:
\begin{align}
\label{eq:loss_total}
\mathcal{L}_{\texttt{total}} = \mathcal{L}_{\texttt{seg}} + \lambda_{\texttt{centroid}} \mathcal{L}_{\texttt{centroid}} + \lambda_{\texttt{hybrid}} \mathcal{L}_{\texttt{hybrid}}.
\end{align}

This multi-component loss ensures that the network simultaneously learns to segment tree canopies, localize centroids, and delineate precise boundaries---addressing the challenges posed by class imbalance and complex spatial distributions in aerial imagery.

\subsection{Hybrid Postprocessing for Final Segmentation}

The final instance segmentation is obtained by fusing the three network outputs---segmentation, centroid, and hybrid SDT-boundary maps---using a multi-stage postprocessing pipeline. This approach leverages complementary cues from each output to refine object boundaries and separate overlapping tree canopies. The pipeline proceeds as follows:

\begin{enumerate}
\item \textbf{Initial Segmentation Refinement:}
The segmentation output (channel 0) is first thresholded using a user-defined threshold to generate a binary mask. Small, spurious regions are removed via a connected-component analysis with a minimum area constraint.

\item \textbf{Hybrid Filtering:}  
The hybrid SDT-boundary map (channel 2), which is expected to contain strong boundary cues with values approaching $-1$, is thresholded to create a binary mask. Regions in the initial binary segmentation that do not exhibit strong boundary signals (i.e., where the hybrid mask is zero) are suppressed, further reducing over-segmentation.

\item \textbf{Centroid Marker Extraction:}  
The centroid map (channel 1) is smoothed using a Gaussian filter to reduce noise. Local maxima are then detected using a peak-finding algorithm with parameters for minimum distance and a minimum intensity threshold. These peaks serve as markers for individual tree instances.

\item \textbf{Watershed Segmentation:}  
With the refined binary segmentation as the mask, the detected centroid markers are labeled and used as seeds for watershed segmentation. The watershed algorithm is applied on the negative of the smoothed centroid map, effectively using it as an “elevation” surface. This process partitions the binary mask into distinct regions corresponding to individual trees.

\item \textbf{Instance Representation (Optional):}  
The watershed labels can be further processed through ellipse fitting or contour extraction to generate vector representations of the detected tree instances.

\end{enumerate}

The integrated postprocessing pipeline thus consolidates the complementary strengths of the segmentation, centroid, and hybrid outputs. By incorporating boundary cues from the hybrid map and spatial markers from the centroid map, the approach produces a final segmentation that achieves improved separation of overlapping canopies and enhanced instance delineation.

\subsection{Dataset Description and Data Splitting}

The dataset used in this study comprises high-resolution aerial imagery provided by the National Land Survey of Finland (NLS), captured during multiple aerial campaigns conducted across Finland. These campaigns, conducted on 17 June 2017, 2 and 7 June 2020, 10 and 15 August 2023, and 7 September 2023, aimed to document tree canopy conditions during the leaf-on season, ensuring optimal visibility for assessing tree health. The imagery includes multispectral bands: red, green, blue, and near-infrared (NIR) which were processed into orthophotos at a spatial resolution of 0.5~m. Orthorectification was performed using a 2~m resolution digital elevation model (DEM), correcting geometric distortions and achieving georeferencing accuracy between 0.5~m and 2~m.

The orthophotos, available through the NLS web interface (National Land Survey of Finland, 2024), are a reliable base for identifying and segmenting tree mortality~\cite{junttila2024significant}. False-color composites, mapping NIR to red, red to green, and green to blue, were utilized alongside standard RGB imagery to enhance visual contrast between healthy and dead trees. This enhancement was particularly beneficial in distinguishing defoliated or entirely brown crowns indicative of dead trees.

Annotations were meticulously performed by two forest health experts using QGIS (version 3.34.11), who manually segmented individual dead tree canopies from the imagery. Two additional experts subsequently verified each annotation to ensure accuracy and consistency. This rigorous annotation process resulted in a dataset comprising 125 aerial images at 25 cm pixel resolution, containing approximately 15,000 centroids, each representing an individual dead tree or canopy segment.

The dataset provides a unique resource for studying forest health, combining high-resolution multispectral imagery, accurate georeferencing, and expert-validated annotations to support advanced machine learning workflows in remote sensing applications (see Table~\ref{tbl:dataset} for a summary of key dataset attributes).

\begin{table}[t!]
    \def\arraystretch{1.1}
    \centering
    \footnotesize
    \caption{Summary of the dataset and its key attributes for model training and evaluation.}
    \begin{adjustbox}{max width=\textwidth}
    \begin{tabular}{lllccccccccc|c}
    \toprule
    ~ & \textsc{\textbf{Feature}} & \textsc{\textbf{Value}} \\
    \midrule

    \multicolumn{3}{l}{\texttt{Dataset}} \\
    & Total Images & 125 \\
    & Total Area Covered & approx. 100 km${}^{2}$ \\ % 98.43
    & Spatial Resolution & 0.25 m/pixel \\
    & Acquisition Dates & 2011, 2013-17, 2019, 2022-23 \\
    & Spectral Bands & Red, Green, Blue, Near-Infrared \\
    & No. of Annotations & $\sim$15,000 individual dead tree polygons \\
    & Avg. Density & $\sim$2 dead trees per hectare \\ % 2.32
    % & Georeferencing Accuracy & 0.5 m to 2 m \\
    \midrule
    \multicolumn{3}{l}{\texttt{Data Partitions}} \\
    & Number of Patches & 87,458 \\
    & Patch Size & $256 \times 256$ \\
    & Patch Overlap (\%) & $\sim$50\% overlapping pixels \\
    % & Data Split & 70\% train, 20\% validation, 10\% test \\
    & Patch Total & 624,966 dead trees \\
    & Dead Tree Count & 454,059 train; 127,411 valid.; 43,496 test \\
    
    \bottomrule
    \end{tabular}
    \end{adjustbox}
\label{tbl:dataset}
\end{table}

\begin{figure*}[t!]
    \centering
    \includegraphics[width=\linewidth]{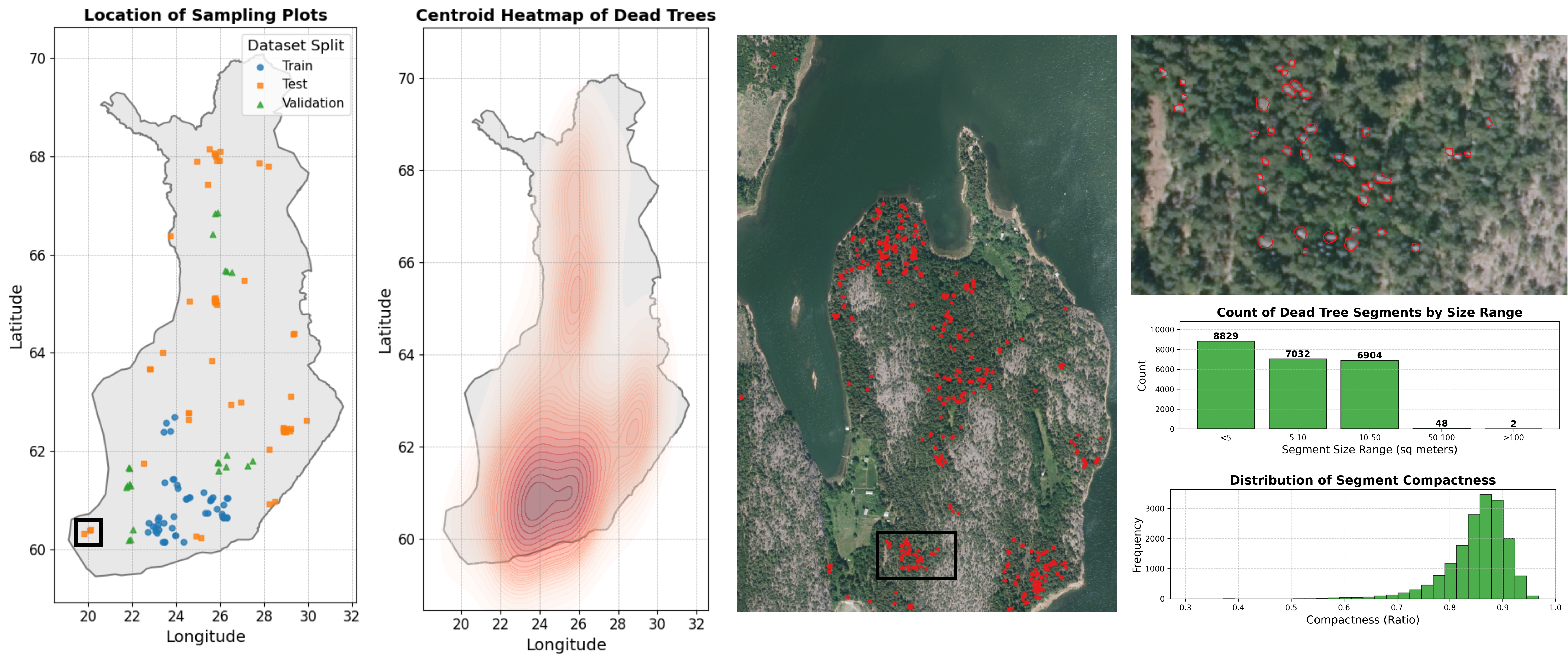}
    \caption{Visualizing the dataset and its spatial characteristics. From left to right: Spatially aware partitioning of the dataset into training (70\%), validation (20\%), and test (10\%) sets, ensuring spatial independence and proportional distribution of dead tree segments across Finland. The partitions are color-coded and overlaid on a map of Finland, with geographic regions clustered based on latitude and longitude bins. Heatmap illustrating the density of dead tree centroids across the dataset, highlighting areas of concentrated tree mortality. A representative aerial image from the dataset, showing a sample plot. A black rectangle on the Finland map indicates the approximate location of this sample plot. A zoomed-in region of the sample plot, with dead tree annotations outlined in red polygons. Binned histogram of dead tree segment sizes, providing insights into the size distribution of mortality segments. The segment compactness distribution depicts the shape characteristics of the detected dead tree regions. Compactness is defined as $\frac{4\pi \times \texttt{area}}{\texttt{perimeter}^2}$, with higher values indicating shapes that are more circular.}
    \label{fig:dataset}
    \vspace{-1em}
\end{figure*}

The dataset was partitioned into training, validation, and test sets in a 70:20:10 ratio using a spatially aware stratification approach. Image patches were grouped into geographic regions based on latitude and longitude bins, with the total number of dead tree segments aggregated for each region. To maintain spatial independence between partitions and reduce the risk of spatial autocorrelation, geographically proximate regions were clustered to ensure that image patches within the same cluster were assigned to a single partition. This method ensured that the distribution of dead tree segments across the training, validation, and test sets was proportional to their overall occurrence, with geographic diversity maintained in each subset. By carefully balancing spatial and annotation characteristics, this strategy provided robust and unbiased partitions for model training and evaluation as shown in Figure~\ref{fig:dataset}.

\subsection{Training and Evaluation Protocol}

The training and evaluation protocol is designed to rigorously assess the multi-task performance of \ours{} in remote sensing instance segmentation. The model is optimized for accurate pixel-level segmentation, precise centroid localization, and detailed boundary delineation via a hybrid SDT-boundary output. These complementary tasks are integrated through a robust postprocessing pipeline that fuses the three outputs into a final segmentation map, which is then evaluated using metrics tailored to capture both pixel-level accuracy and instance-level detection quality.

\mypar{Evaluation metrics.} To evaluate the model's performance, we employ a combination of pixel-level and instance-level metrics that provide complementary insights into segmentation quality and localization accuracy.

\begin{enumerate}
    \item \textbf{Pixel IoU (Intersection over Union):}
    Pixel IoU measures the overlap between the final predicted segmentation mask and the ground truth at the pixel level. For a binary segmentation task, IoU is defined as:
    \begin{align}
    \label{eq:iou}
    \texttt{IoU} = \frac{\sum_{i} P_i \cdot G_i}{\sum_{i} (P_i + G_i - P_i \cdot G_i)},
    \end{align}
    where $P_i$ and $G_i$ denote the predicted and ground truth binary labels for pixel $i$, respectively. Higher IoU values indicate better segmentation accuracy.

    \item \textbf{Tree IoU:}
    Tree IoU extends pixel-level IoU to evaluate the segmentation quality at the instance level. Let $T_p$ and $T_g$ represent the sets of predicted and ground truth tree instances, respectively. The Tree IoU is computed as:
    \begin{align}
    \label{eq:tree_iou}
    \texttt{Tree IoU} = \frac{\sum_{k} \texttt{IoU}(T_p^k, T_g^k)}{|T_g|},
    \end{align}
    where $\texttt{IoU}(T_p^k, T_g^k)$ is the IoU of the $k$-th predicted instance with its corresponding ground truth instance. This metric is particularly relevant in ecological applications as it accounts for overlapping boundaries and densely clustered tree regions.

    \item \textbf{Centroid Localization Error:}
    Centroid localization error is quantified using the Root Mean Squared Error (RMSE) between the predicted centroids and the ground truth centroids:
    \begin{align}
    \label{eq:rmse}
    \texttt{RMSE} = \sqrt{\frac{1}{N} \sum_{i=1}^{N} \left[ (x_i^p - x_i^g)^2 + (y_i^p - y_i^g)^2 \right]},
    \end{align}
    where $(x_i^p, y_i^p)$ and $(x_i^g, y_i^g)$ are the predicted and ground truth centroid coordinates for the $i$-th instance, respectively. This metric captures the accuracy of instance localization, which is crucial for downstream applications such as biomass estimation.

    \item \textbf{Precision, Recall, and F1-Score:} At the instance level, we define true positives (TP) as the number of predicted instances that match a ground truth instance (with an Intersection over Union above a set threshold), false positives (FP) as predictions that do not match any ground truth, and false negatives (FN) as ground truth instances that are missed by the model. These metrics are computed as:
    \begin{align}
    \label{eq:metrics}
    \texttt{Precision} &= \frac{\texttt{TP}}{\texttt{TP} + \texttt{FP}},\\[1ex]
    \texttt{Recall} &= \frac{\texttt{TP}}{\texttt{TP} + \texttt{FN}},\\[1ex]
    \texttt{F1-Score} &= 2 \cdot \frac{\texttt{Precision} \cdot \texttt{Recall}}{\texttt{Precision} + \texttt{Recall}}.
    \end{align}
    These measures assess the model’s ability to correctly detect individual tree instances while minimizing spurious detections.
\end{enumerate}

\mypar{Robustness and generalization.} The protocol includes repeated experiments across five random dataset splits to ensure robust evaluation. Metrics are reported as mean values with 95\% confidence intervals. Furthermore, generalization is assessed by testing the model on unseen geographic regions, thereby demonstrating its adaptability across diverse spatial and spectral domains.

The proposed methodology integrates the multi-task \ours{} architecture, a hybrid loss function, and a comprehensive postprocessing pipeline to overcome challenges in aerial imagery instance segmentation, such as overlapping objects and spectral variability. The final segmentation, which is derived by fusing the segmentation, centroid, and hybrid outputs, is evaluated using the aforementioned metrics to quantify the framework's effectiveness and robustness rigorously.

\section{Experimental Results}
\label{sec:results}

The experimental evaluation highlights the effectiveness of the proposed hybrid postprocessing pipeline for standing dead tree segmentation and centroid localization. Results are analyzed through quantitative metrics, statistical significance, and ablation studies, demonstrating improvements in pixel-level and instance-level segmentation accuracy. This section also discusses the practical implications of these improvements for ecological monitoring tasks.

\mypar{Quantitative results.} The proposed \ours{}, which integrates a hybrid postprocessing pipeline to fuse segmentation, centroid, and boundary predictions, was evaluated against multiple models including U-Net as the baseline model. As summarized in Table~\ref{tbl:metrics}, \ours{} achieves notable improvements in key metrics. In terms of pixel-level accuracy, the proposed model attains a Mean IoU of 0.2588 compared to 0.1823 for U-Net. At the instance level, the Mean Tree IoU increases markedly from 0.2622 (U-Net) to 0.3710 (\ours{}). Furthermore, while the precision of \ours{} is comparable to the baseline (0.6418 vs. 0.6742), its recall improves substantially (0.6687 vs. 0.4667), resulting in an enhanced F1-score (0.5895 vs. 0.4471). Additionally, the centroid localization error is reduced to 3.70 pixels, indicating more accurate detection of tree instances. These results underscore the pipeline’s ability to refine instance boundaries and recover missed tree instances, making it particularly effective for applications where high recall is critical.

\begin{table*}[t!]
    \def\arraystretch{1.1}
    \centering
    \footnotesize
    \caption{Achieved standing dead tree segmentation results by the baseline and proposed Dual-Task Learning approach with \ours{}. The best performer is marked by \textbf{bold text}; the second best is \underline{underlined}.}
    \begin{adjustbox}{max width=\textwidth}
    \begin{tabular}{lccccccccccc|c}
    \toprule
     \textsc{\textbf{Method}} & \textsc{\textbf{Mean IoU Pixels}} & \textsc{\textbf{Mean IoU Trees}} & \textsc{\textbf{Precision}} & \textsc{\textbf{Recall}} & \textsc{\textbf{F1-Score}} & \textsc{\textbf{Centroid Error (px)}} \\
    \midrule

    U-Net~\cite{ronneberger2015u} & 0.1823 & 0.2622 & 0.6742 & 0.4667 & 0.4471 & 8.60 \\

    KokoNet~\cite{junttila2024significant} & 0.0817 & 0.2158 & 0.6557 & 0.3930 & 0.3864 & 8.89 \\

    DETR~\cite{carion2020end} & 0.1546 & 0.2479 & \underline{0.7098} & 0.4403 & 0.4545 & 7.50 \\

    % Self-Attention U-Net & 0.1731 & 0.2752 & 0.6407 & 0.4801 & 0.4753 & 6.09 \\

    DINOv2~\cite{oquab2023dinov2} & 0.1781 & 0.2722 & 0.6663 & 0.5349 & 0.5130 & 4.74 \\

    BEiT~\cite{bao2021beit} & 0.1783 & 0.2817 & 0.7207 & 0.4784 & 0.5021 & 7.37 \\

    PSPNet~\cite{zhao2017pyramid} & 0.2093 & 0.3234 & 0.6372 & 0.6463 & 0.5703 & 3.12 \\

    FPN~\cite{lin2017feature} & 0.2147 & 0.3251 & \textbf{0.7198} & 0.5868 & 0.5770 & 4.15 \\

    DeepLabv3+~\cite{chen2018encoder} & 0.2288 & 0.3263 & 0.6657 & 0.6015 & 0.5691 & 4.77 \\

    HCF-Net~\cite{xu2024hcf} & 0.2365 & 0.3336 & 0.6903 & 0.5993 & 0.5501 & 4.36 \\

    MaskFormer~\cite{cheng2021per} & 0.2397 & \underline{0.3503} & 0.6202 & \textbf{0.6882} & 0.5774 & \underline{3.04} \\
    
    U-Net (EfficientNet-B7)~\cite{ronneberger2015u} & \underline{0.2428} & 0.3420 & 0.6319 & 0.6684 & 0.5619 & \textbf{2.71} \\

    UNet++ (EfficientNet-B7)~\cite{zhou2018unet++} & \underline{0.2428} & 0.3358 & 0.6347 & 0.6565 & \underline{0.5803} & 3.30 \\

    % U-Net with ResNet34~\cite{ronneberger2015u} & 0.2608 & 0.3669 & 0.6434 & 0.6694 & 0.5916 & 3.73 \\
    
    % TreeMort-1T-UNet    & 0.2446 & 0.3398 & 0.6336 & 0.6316 & 0.5705 & 3.15 \\
    \rowcolor{lncolor}
    \ours{} & \textbf{0.2588} & \textbf{0.3710} & 0.6418 & \underline{0.6687} & \textbf{0.5895} & 3.70 \\
    
    \bottomrule
    \end{tabular}
    \end{adjustbox}
\label{tbl:metrics}
\end{table*}

% \begin{figure}
%     \centering
%     \includegraphics[width=1.0\linewidth]{metrics.png}
%     \caption{Bar plots comparing the performance of models in terms of Mean Pixel IoU and Mean Tree IoU, with error bars indicating the standard deviation.}
%     \label{fig:metrics}
% \end{figure}

\begin{figure*}[!ht]
    \centering
    \subfloat[Ground Truth]{\includegraphics[width=0.32\linewidth]{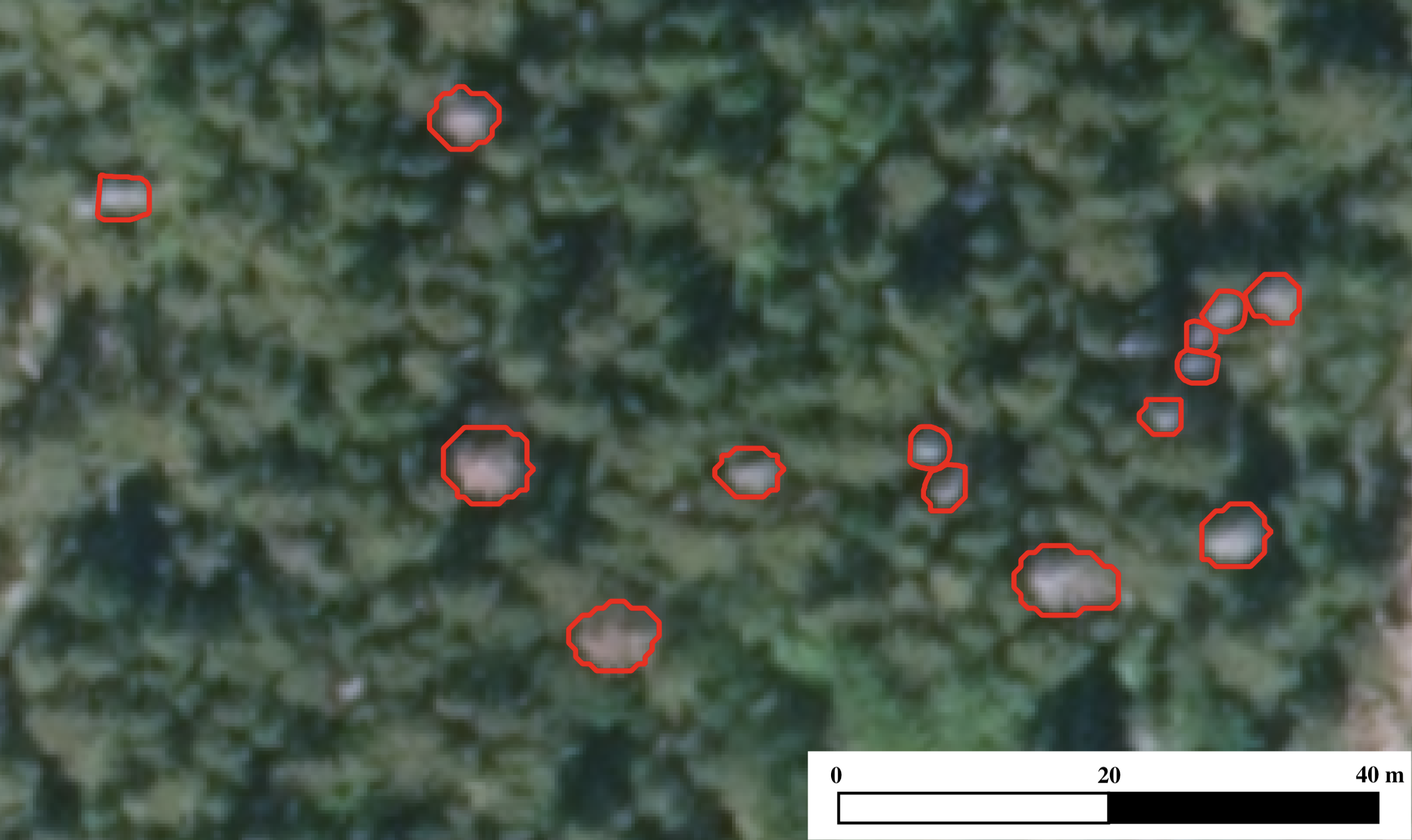} \label{subfig:input}}
    \hfill
    \subfloat[Model Prediction]{\includegraphics[width=0.32\linewidth]{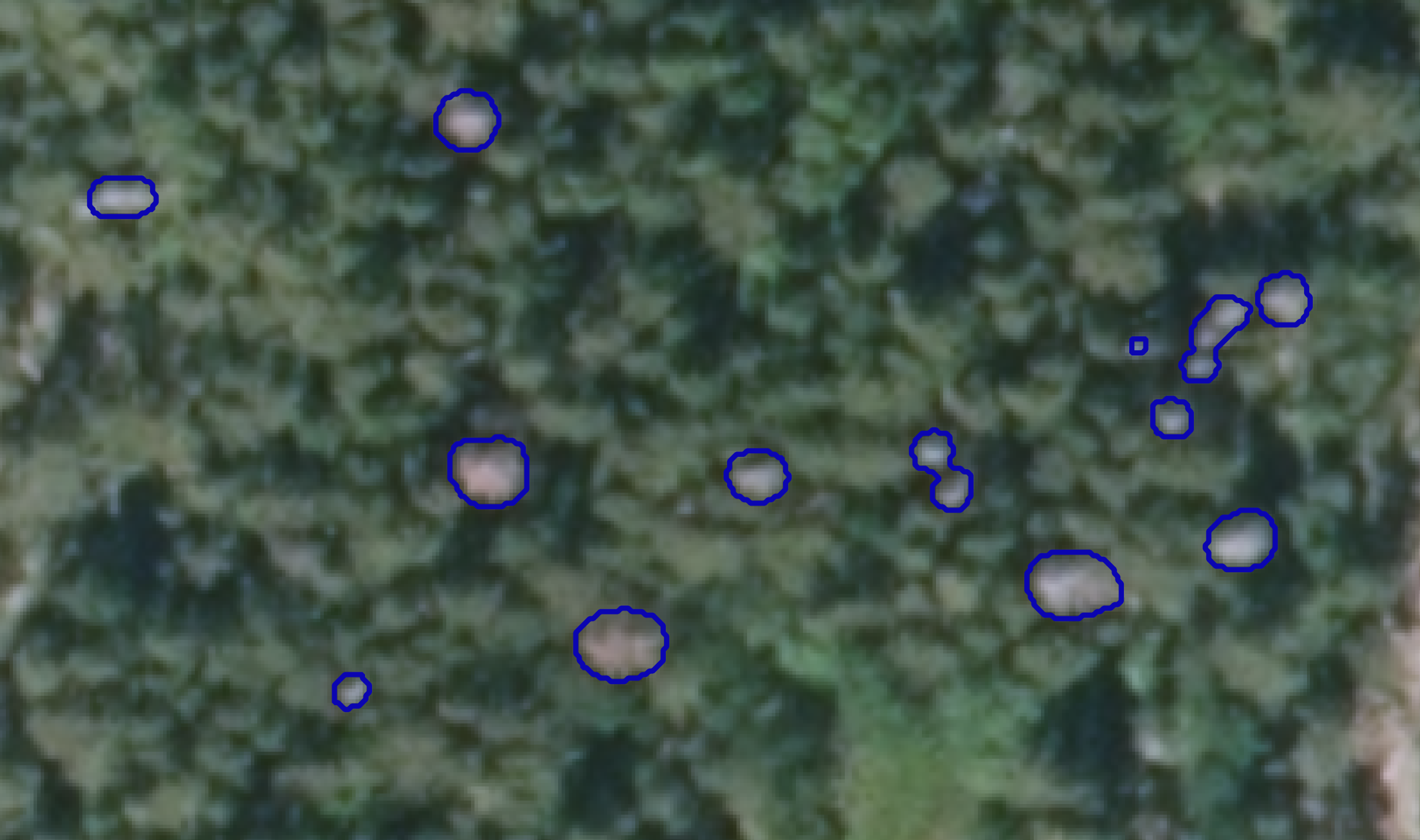} \label{subfig:prediction}}
    \hfill
    \subfloat[Postprocessed Output]{\includegraphics[width=0.32\linewidth]{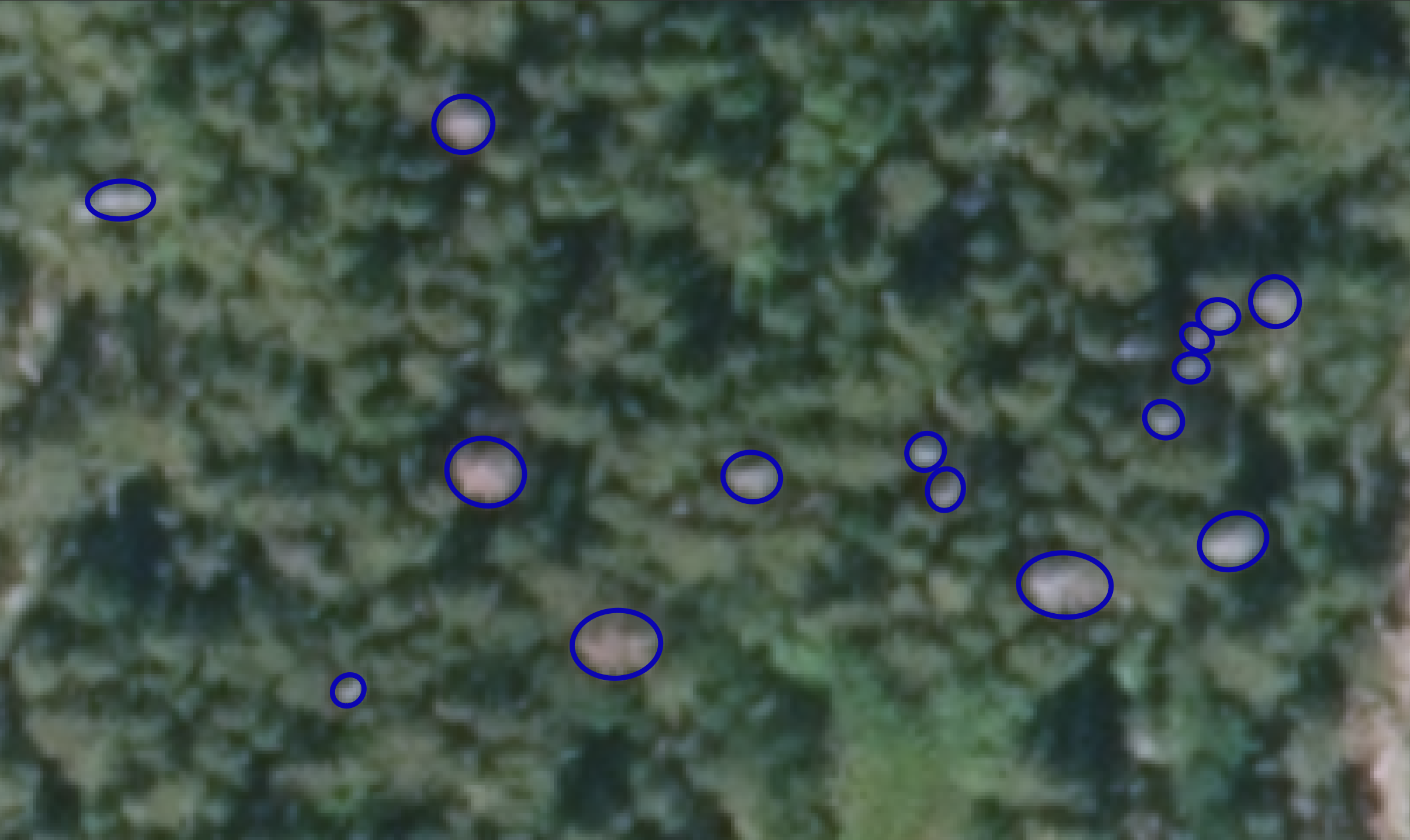} \label{subfig:postprocessed}}
    \caption{Sample images showcasing the segmentation process. (a) Input aerial image with ground truth annotation, (b) Model’s initial baseline prediction, and (c) Final postprocessed segmentation.}
    \label{fig:samples}
\end{figure*}

\mypar{Statistical analysis.} The statistical significance of the observed improvements was evaluated using paired tests as presented in Table~\ref{tbl:significance}. For Pixel IoU, the p-value of $<0.001$ indicated that the difference between U-Net and \ours{} results is statistically significant. Similarly, for Tree IoU, the p-value of $<0.001$ confirmed a significant improvement. The confidence intervals (CIs) further supported this conclusion; while the U-Net and \ours{} CIs for Pixel IoU overlap, those for Tree IoU were distinct, demonstrating meaningful gains in instance-level accuracy (Figure~\ref{fig:samples}). These results highlight the pipeline's significant contribution to instance-level segmentation, particularly in applications where individual tree delineation is critical.

\begin{table*}[t!]
    \def\arraystretch{1.1}
    \centering
    \footnotesize
    \caption{Statistical significance analysis of model performance metrics across partitions.}
    \begin{adjustbox}{max width=\textwidth}
    \begin{tabular}{lccccccccccc|c}
    \toprule
     \textsc{\textbf{Metric}} & \textsc{\textbf{p-value}} & \textsc{\textbf{U-Net}} & \textsc{\textbf{\ours{}}} \\
    \midrule

    Mean IoU Pixels & 1.07 $\times 10^{-4}$ & (0.1381, 0.2266) & (0.2226, 0.2949) \\
    Mean IoU Trees  & 7.91 $\times 10^{-5}$ & (0.2020, 0.3224) & (0.3199, 0.4222) \\

    \bottomrule
    \end{tabular}
    \end{adjustbox}
\label{tbl:significance}
\end{table*}

\mypar{Ablation studies.} The results of the ablation study, summarized in Table~\ref{tbl:abulation}, demonstrate the incremental contributions of the simplified postprocessing pipeline components to segmentation performance. The baseline configuration, utilizing initial \ours{} outputs without postprocessing, achieves a Mean IoU of 0.2347 for pixels and 0.3299 for dead trees, with a centroid localization error of 4.11 pixels. Incorporating segment filtering, which removes small or irregular artifacts, yields modest improvements in Mean IoU for pixels (0.2362). However, centroid localization error increases slightly to 4.13 pixels, likely due to excluding marginally valid segments. In contrast, the addition of watershed segmentation significantly improves pixel and tree IoUs to 0.2620 and 0.3689, respectively, while reducing centroid error to 3.78 pixels by effectively delineating overlapping tree canopies. Further gains were achieved when both components were combined in the entire postprocessing pipeline. Mean IoU metrics rose to 0.2588 (pixels) and 0.3710 (trees), alongside an improved centroid localization error of 3.70 pixels. These results underline the importance of leveraging spatial consistency through watershed segmentation and artifact removal via filtering to enhance segmentation quality, particularly for complex tree canopy structures in high-resolution aerial imagery.

\begin{table*}[t!]
    \def\arraystretch{1.1}
    \centering
    \footnotesize
    \caption{Ablation study results highlight individual components' impact on model performance.}
    \begin{adjustbox}{max width=\textwidth}
    \begin{tabular}{lccccccccccc|c}
    \toprule
     \textsc{\textbf{Method}} & \textsc{\textbf{Mean IoU}} & \textsc{\textbf{Mean IoU}} & \textsc{\textbf{Centroid}} \\
     ~ & \textsc{\textbf{Pixels}} & \textsc{\textbf{Trees}} & \textsc{\textbf{Error (px)}} \\
    \midrule

    Raw Segments             & 0.2347 & 0.3299 & 4.11 \\
    Segment Filtering        & 0.2362 & 0.3299 & 4.13 \\
    Watershed Segmentation   & 0.2620 & 0.3689 & 3.78\\
    Final Segmentation       & 0.2588 & 0.3710 & 3.70 \\
    
    \bottomrule
    \end{tabular}
    \end{adjustbox}
\label{tbl:abulation}
\end{table*}

\mypar{Practical implications.} The improvements achieved by the hybrid postprocessing pipeline underscore its potential for ecological monitoring tasks such as tree mortality mapping and forest health assessments. The significant increase in Tree IoU suggests that the pipeline excels at delineating individual trees, even in complex scenarios with overlapping or sparse instances. However, the observed increase in false positives indicates a trade-off, suggesting that further refinement is needed to mitigate over-segmentation while maintaining high instance delineation performance.

\section{Discussion}
\label{sec:discussion}

This study presents a robust and efficient pipeline for instance-level segmentation of dead trees using aerial imagery. Integrating the hybrid postprocessing methodology and a spatially stratified dataset partitioning has demonstrated significant improvements in segmentation performance. Notably, the watershed segmentation step substantially improved both pixel- and tree-level IoU metrics while reducing centroid localization errors, as illustrated in Table~\ref{tbl:metrics}. This improvement underscores the effectiveness of utilizing spatial and intensity-based markers to refine segmentation boundaries.

\begin{figure}[!t]
    \centering
    \includegraphics[width=\linewidth]{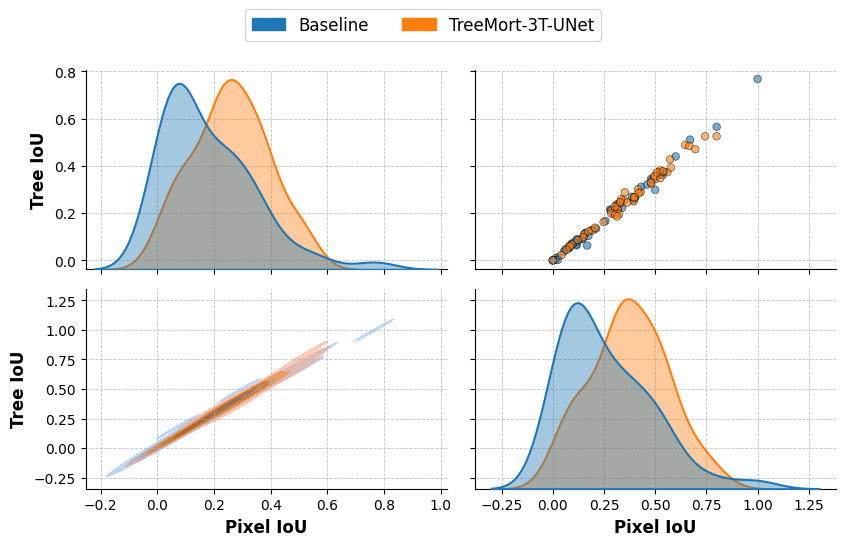}
    \caption{Paired scatter plot comparing Mean Pixel IoU and Mean Tree IoU across different model configurations. Each data point represents a specific model's performance, highlighting the relationship and potential trade-offs between pixel-level accuracy and instance-level segmentation precision.}
    \label{fig:paired}
    \vspace{-1em}
\end{figure}

The simplified postprocessing pipeline, which removed the refinement module and graph partitioning, retained competitive performance while reducing computational complexity. The ablation study results highlight the incremental contributions of each step, with watershed segmentation emerging as the most impactful component. The paired IoU plot (Figure~\ref{fig:paired}) further illustrates the balance between pixel-level and tree-level segmentation accuracy achieved through the pipeline.

Despite these advancements, the methodology is not without limitations. The reliance on manual annotations introduces potential biases, and the model’s performance in dense forest regions with overlapping canopies remains challenging. Additionally, although geographically diverse, the dataset is limited to Finland, potentially restricting the generalizability of the findings to other regions with varying vegetation types and environmental conditions.

The implications of this work extend beyond dead tree detection. The proposed pipeline can be adapted for broader ecological monitoring tasks, such as identifying tree species or mapping vegetation health. Future efforts could explore the integration of hyperspectral imagery, advanced neural architectures, and more automated annotation techniques to enhance scalability and accuracy. This work lays the groundwork for operational applications in precision forestry and conservation by addressing the outlined limitations.

\section{Conclusions}
\label{sec:conclusions}

This study advanced tree segmentation in aerial imagery through a hybrid postprocessing framework that refines instance boundaries and reduces over-segmentation errors. Evaluated on boreal forest datasets, the approach achieved a 41.5\% improvement in Mean Tree IoU and a 57\% reduction in Centroid Error, demonstrating its efficacy in resolving ambiguities in dense canopies. The framework balances precision-recall trade-offs by harmonizing watershed-based refinement with adaptive filtering, offering a scalable solution for large-scale forest monitoring.

Future work will prioritize wall-to-wall tree mortality prediction across expansive regions, leveraging aerial and satellite imagery and distributed UAV fleets to map dead tree distributions at landscape scales. Such capabilities could directly benefit downstream applications, including wildfire risk assessment (identifying fuel hotspots), carbon stock quantification (tracking emissions from decaying biomass), and biodiversity conservation (mapping habitat fragmentation). Integrating dead tree counts with geospatial analytics could also inform precision forestry operations, such as targeted salvage logging and reforestation planning. By advancing scalable segmentation tools, this work lays the groundwork for actionable insights in ecological management, bridging the gap between remote sensing innovation and on-the-ground conservation efforts.

% \clearpage
% \input{sec/ack}
{
    \small
    \bibliographystyle{ieeenat_fullname}
    \bibliography{main}
}

\end{document}